\documentclass[conference]{IEEEtran}
\IEEEoverridecommandlockouts
\usepackage{cite}
\usepackage{amsmath,amssymb,amsfonts}
\usepackage{algorithmic}
\usepackage{graphicx}
\usepackage{booktabs}
\usepackage{graphicx}
\usepackage{textcomp}
\usepackage{xcolor}
\usepackage{soul}
\usepackage{float}
\usepackage{multirow} 
\usepackage[hidelinks]{hyperref}
\usepackage{subcaption}
\def\BibTeX{{\rm B\kern-.05em{\sc i\kern-.025em b}\kern-.08em
    T\kern-.1667em\lower.7ex\hbox{E}\kern-.125emX}}
\begin{document}

\title{Denoising-Aware Inversion: Revealing Privacy Risks in Noise-Protected Text Embeddings}

\author{
\IEEEauthorblockN{
Yubo Wang$^{1}$,
Shujie Cui$^{1}$,
James Bailey$^{1}$,
Hongzhi Yin$^{2}$,\\
Wenyu Liang$^{1}$,
Min Tang$^{1}$,
Shiyue Qin$^{3}$,
Weiqing Wang$^{1}$
}

\IEEEauthorblockA{
$^{1}$Monash University, Melbourne, Australia\\
$^{2}$The University of Queensland, Brisbane, Australia\\
$^{3}$Northeastern University, Shenyang, China\\
\{yubo.wang2, shujie.cui, james.a.bailey, Teresa.Wang\}@monash.edu\\
wlia0047@student.monash.edu,
bupttm@163.com,
h.yin1@uq.edu.au,
qinsy@mail.neu.edu.cn
}
}

\maketitle

\begin{abstract}
Dense text embeddings are widely used in data mining, retrieval, and downstream machine learning systems due to their compact and semantically rich representations, but recent embedding inversion attacks have shown that they can expose substantial information about the original text, leading to serious privacy leakage risks. A common defense is to release perturbed embeddings by adding Gaussian noise, which is simple yet effective against standard inversion attacks and does not significantly degrade embedding utility for downstream tasks. However, it remains unclear whether such noise-protected embeddings are sufficiently safe against adaptive attackers that explicitly account for the perturbation process. 
In this paper, we study text embedding inversion in a noise-protected setting, where the attacker can observe only noisy embeddings and has no access to clean embedding targets. We first analyze why existing generative inversion methods fail under this setting and identify a ``Double Noise Trap'', which fundamentally prevents standard generative inversion models from achieving high-quality reconstruction. To address this challenge, we propose DAEI, a denoising-aware embedding inversion pipeline that combines a residual denoising autoencoder with generative text inversion where the denoiser is trained in an unsupervised manner using Stein's unbiased risk estimate to enable denoising from noisy observations alone. 
Extensive experiments show that DAEI achieves approximately 154\% relative improvement in BLEU over the existing generative inversion baseline, while also improving token-level F1 and ROUGE-L by 32--60\%. The promising inversion performance of DAEI challenges the prevailing assumption that simple Gaussian perturbation is sufficient to prevent sensitive information leakage from embedding representations. The code for reproduction is provided below: https://github.com/ChrisWang233/DAEI

\end{abstract}


\section{Introduction}

Dense text embeddings have become a standard data representation in modern data mining and machine learning systems \cite{reimers2019sentence,karpukhin2020dense,muennighoff2023mteb}. Across applications such as semantic search, user modeling, vector databases, and retrieval-augmented analytics, raw textual records are increasingly transformed into compact continuous vectors that are stored, searched, or used for various tasks by downstream systems \cite{lewis2020retrieval,zhao2024dense,gao2023retrieval}. This representation is attractive because it transforms raw data into low-dimensional and semantically rich vector representations, making subsequent tasks more efficient\cite{reimers2019sentence,zhao2024dense}.
Yet these benefits create a potential privacy issue: high-quality embeddings are explicitly trained to preserve semantic and lexical information, and may therefore also preserve sensitive entities and attributes\cite{song2020information,li2023sentence}. Consequently, any leakage allows adversaries to apply sophisticated extraction techniques to recover the underlying sensitive data, resulting in a  privacy vulnerability.

Recent works have made this privacy risk concrete. Generative embedding inversion attacks reconstruct full sentences from sentence embeddings, rather than simply recovering unordered keywords \cite{li2023sentence}. Vec2Text further frames inversion as controlled generation by introducing an iterative correction framework, substantially improving reconstruction accuracy for short inputs \cite{morris2023text}. In practice, embeddings may be exposed through compromised vector databases, honest-but-curious storage or retrieval providers, and distributed  systems that transmit embeddings online across components where the adversaries can attack the embedding during transmission \cite{luo2025prompt, wang2025privacy}. Once an adversary can observe these embeddings, the embeddings themselves may become a text disclosure channel.

One natural defense strategy is to perturb embeddings before release. Differential privacy formalizes the broader principle of adding calibrated noise to sensitive computations \cite{dwork2006calibrating}, and prior work on privacy-preserving textual analysis has shown that perturbations in embedding spaces can maintain downstream utility while reducing disclosure risks in data mining workflows \cite{feyisetan2020privacy,xu2020differentially,du2023sanitizing}.
The most prevalent defense mechanism is additive Gaussian noise injection \cite{morris2023text,zhuang2024understanding,seputis2025rethinking}. In practice, adding zero-mean Gaussian noise with standard deviation $\sigma=0.01$ to the final representations of the target embedder can severely degrade generative inversion performance while barely affecting downstream task utility \cite{morris2023text}. 
However, as far as we know, there are no existing embedding inversion methods studying whether embedding-to-text inversion remains possible under this noise-protected setting. The closest related work is ZSInvert \cite{zhang2025universal}, which focuses on zero-shot inversion instead of focusing on noise-protected setting. We consider it the closest work because it includes an experimental case study to show that the proposed approach is noise robust \cite{zhang2025universal}. The possible reason is that its search-based optimization offers some robustness to noisy embeddings, but the recovered texts are often less readable and weakly token-aligned, which is also validated in our experiments.
Therefore, we include it as a noise-robust baseline, while noting that conventional generative inversion methods remain largely ineffective under the noise-protected embedding setting.

This difficulty goes beyond a harder inversion task: noise creates a fundamentally different and underexplored inversion paradigm.
In this setting, adversaries are deprived of clean embeddings and can observe only noisy vectors. This makes inversion substantially more challenging for two reasons: first, the absence of clean targets invalidates standard supervised denoising objectives that rely on reconstruction error against clean references; second, the inverse mapping must disentangle the underlying semantic signal from high-dimensional noise before text can be reliably recovered. 
Consequently, these constraints motivate the central question of our work: can adversaries recover sufficient textual information from black-box noisy embeddings without relying on clean embeddings?

We investigate this question from a security-evaluation perspective. Specifically, we study the no-clean-embedding regime and propose \textbf{DAEI}: \textbf{D}enoising-\textbf{A}ware \textbf{E}mbedding \textbf{I}nversion, an integrated pipeline for measuring the privacy risk of noise-protected text embeddings.
Building upon Vec2Text as our generative backbone for text recovery, we prepend a residual denoising autoencoder (DAE) \cite{vincent2008extracting} to the inversion decoder to separate semantic signals from high-dimensional noise. A conventional DAE typically relies on a Mean Squared Error (MSE) objective against clean reference embeddings, which are unavailable in our noisy-only setting \cite{vincent2008extracting,vincent2010stacked}. Under the Gaussian noise model, we therefore use Stein's unbiased risk estimate (SURE) \cite{stein1981estimation,ramani2008monte} to remove this reliance on clean embeddings, yielding an unbiased risk estimate computable from noisy observations alone. We further use Monte-Carlo Hutchinson estimation \cite{ramani2008monte} to approximate the required divergence term with an explicit variance-based error bound.
To align denoising with text generation, we jointly fine-tune both modules. This coupled optimization guides the DAE to preserve critical semantic directions, while dynamically adapting the inverter to the refined embedding space.
Furthermore, unlike methods such as Noise2Noise \cite{lehtinen2018noise2noise} that require multiple noisy observations of each sample which are usually unavailable in real world applications, \textsc{DAEI} operates with only one noisy observation per sample.

In summary, our main contributions are as follows:
\begin{itemize}

    \item We challenge the prevailing assumption that noise guarantees text embedding privacy. We are the first to formulate and systematically evaluate an adaptive ``denoising-inversion" approach. By demonstrating that underlying textual information can still be reliably recovered from noise-protected embeddings, we provide a critical re-evaluation of current representation security standards.
    
    \item We systematically analyze the failure of existing generative inversion approach in reconstructing text from noisy embeddings and introduce \textbf{``Double Noise Trap"} (Section \ref{sec:dn_trap}).
    
    \item To the best of our knowledge, we are the first to study denoising autoencoders for noisy text embedding inversion under a strict no-clean-target setting.

    \item We propose \textsc{DAEI}, a novel denoising-aware inversion pipeline that couples DAE pre-training, DAE-shielded inversion, components joint fine-tuning, and DAE-shielded correction, enabling robust text recovery from noisy observations.

    \item We validate \textsc{DAEI} across multiple benchmarks and embedding backbones. Compared with noisy Vec2Text, \textsc{DAEI} achieves approximately 154\% relative improvement in BLEU and 32--60\% gains in F1 and ROUGE-L, while also substantially outperforming ZSInvert. 

\end{itemize}

\section{Problem Formulation and Motivation}
\label{sec:problem}

\subsection{Problem Formulation}
\label{sec:problem_task}

\begin{figure}[t]
    \centering
    \includegraphics[page=1,width=\linewidth,
    trim=180 160 220 100,
    clip]{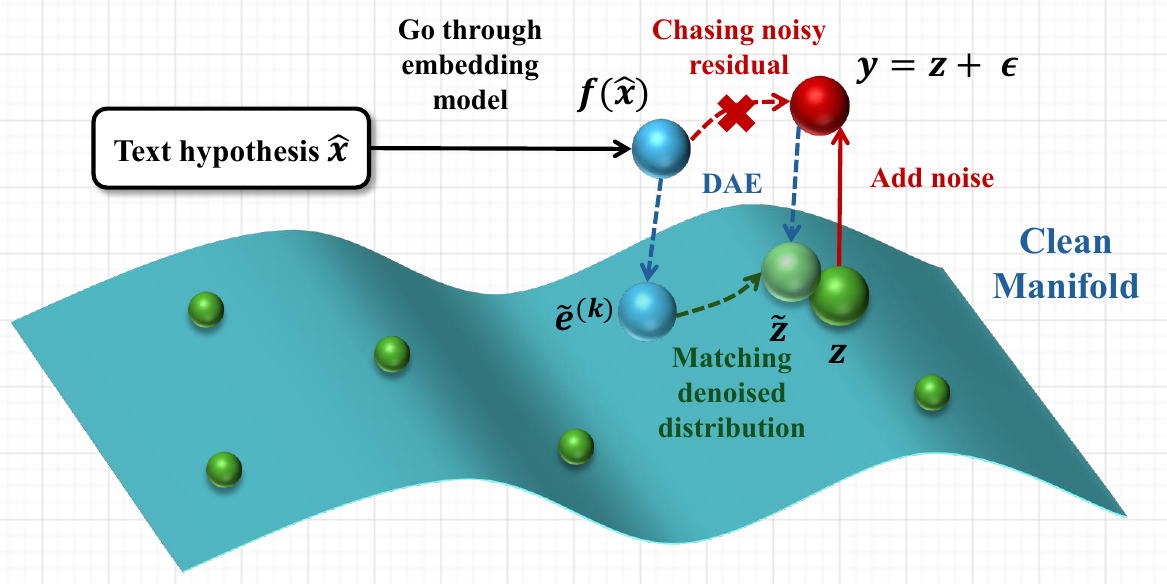}
    \caption{
        Geometric intuition of the ``Double Noise Trap''.
        Additive Gaussian noise pushes the target embedding
        off the clean text manifold, causing vanilla inversion
        to chase off-manifold directions.
        \textsc{DAEI} instead projects the noisy representation
        back toward the clean manifold before inversion.
        }    
    \label{fig:dn_trap}
\end{figure}

Let $x \in \mathcal{X}$ denote an input text and let 
$f_{\mathrm{clean}}: \mathcal{X} \rightarrow \mathbb{R}^{d}$ be a frozen black-box text encoder. In the standard setting, the exact latent representation is given by
\begin{equation}
    z = f_{\mathrm{clean}}(x),
\end{equation}
We study a privacy-preserving regime where the system releases only a perturbed version of the final embedding, denoted as $y$:
\begin{equation}
    y = f(x) = z + \epsilon, \qquad
    \epsilon \sim \mathcal{N}(0,\sigma^2 I_d).
    \label{eq:noisy_embedding}
\end{equation}
where $f$ denotes a noisy embedder, and the standard deviation of noise scale $\sigma$ is assumed to be known or estimable, but the clean embedding $z$ remains strictly inaccessible to the attacker. 

\textbf{Adversary assumption:} In our threat model, we assume that the adversary has no knowledge of the embedder such as architecture and parameters. 1) Adversary can only query the released embedding interface as a black box and obtain an output embedding for a given attacker-chosen input text, thereby constructing noisy embedding–text training pairs $(y,x)$; 2) The adversary is assumed to know that the perturbation is additive Gaussian noise, since it is a standard and widely adopted mechanism for protecting text embedding; 3) The adversary either knows the noise scale $\sigma$, as prior work commonly adopts $\sigma=0.01$ to achieve the best tradeoff between effectiveness and privacy \cite{morris2023text}; or can estimate $\sigma$ through repeated calibration queries on any adversary-chosen input text by observing the variation of the returned embeddings. Specifically, for $n$ repeated observations of the same calibration text,
$\{y^{(j)}=z+\epsilon^{(j)}\}_{j=1}^{n}$, an unbiased estimator of the noise variance is
\begin{equation}
    \hat{\sigma}^{2}
    =
    \frac{1}{d(n-1)}
    \sum_{j=1}^{n}
    \|y^{(j)}-\bar{y}\|_2^2,
    \qquad
    \bar{y}=\frac{1}{n}\sum_{j=1}^{n}y^{(j)} .
\end{equation}
Under Gaussian noise, $d(n-1)\hat{\sigma}^2/\sigma^2$ follows a chi-square distribution with $d(n-1)$ degrees of freedom \cite{casella2024statistical}, so the relative standard error of $\hat{\sigma}^2$ is $\sqrt{2/(d(n-1))}$. For a 768-dimensional embedder, even $n=10$ calibration queries give a relative standard error of about $1.7\%$ for $\hat{\sigma}^2$ and about $0.85\%$ for $\hat{\sigma}$. 
Please note that, even though an attacker can probably estimate the clean embedding for the text they input to the encoder multiple times and these clean embeddings might be helpful in reconstructing the original text \textbf{that the adversaries input}. But the text we want to protect is \textbf{NOT} the text that the adversaries input but the hidden text of the target systems. For these protected texts, the adversaries only observe the released noisy embeddings and they cannot estimate the clean embeddings for them. 

Given the observed noisy embedding $y$ and black-box embedder $f$, the task is to reconstruct the original text $x$. A natural initial attempt is to directly train an inversion model on the noisy pairs $(y,x)$.  However, as observed in prior \textbf{experiments} by Zhuang et al. \cite{zhuang2024understanding}, this strategy remains largely insufficient. Below, we analyze \textbf{theoretically} why this naive approach fails.

\subsection{Vec2Text Fails due to ``Double Noise Trap"}
\label{sec:dn_trap}

Generative inversion methods such as Vec2Text \cite{morris2023text} are designed to map a target embedding $z$ back to text $x$ by modeling $p_{\theta}(x \mid z)$ via an inverter, and refine a hypothesis $\hat{x}$ via a corrector by comparing the target embedding with the embedding of the hypothesis $f(\hat{x})$. Under a noisy embedding interface, both quantities are observed with perturbations:
\begin{equation}
    y_{\mathrm{tar}} = z + \epsilon_1,
    \qquad
    y_{\mathrm{hyp}} = f(\hat{x}) = f_{\mathrm{clean}}(\hat{x}) + \epsilon_2,
\end{equation}

where $\epsilon_1,\epsilon_2\sim\mathcal{N}(0,\sigma^2I_d)$ are independent. The observed correction residual is therefore
\begin{equation}
\begin{aligned}
    \Delta_{\mathrm{obs}}
    &=
    y_{\mathrm{tar}} - y_{\mathrm{hyp}} \\
    &=
    \underbrace{z- f_{\mathrm{clean}}(\hat{x})}_{\text{semantic residual}}
    +
    \underbrace{(\epsilon_1-\epsilon_2)}_{\text{noise residual}} .
\end{aligned}
\label{eq:double_noise_residual}
\end{equation}

Thus, the correction signal contains not only the desired semantic residual, but also a residual noise term formed by subtracting two independent perturbations. Since $\epsilon_1-\epsilon_2\sim\mathcal{N}(0,2\sigma^2I_d)$, the residual noise has doubled variance, and its typical magnitude is
\begin{equation}
    \|\epsilon_1-\epsilon_2\|_2 \approx \sqrt{2}\sigma\sqrt{d}.
    \label{eq:double_noise_scale}
\end{equation}
For a 768-dimensional embedding with $\sigma=0.01$, this residual noise magnitude is approximately $0.39$, which is larger than the single-noise displacement $\sigma\sqrt{d}\approx0.28$.

As shown in Fig. \ref{fig:dn_trap}, clean embeddings reside on a text-induced manifold, and additive noise pushes $y$ off this manifold. Consequently, the corrector attempting to match the hypothetical embedding $f(\hat{x})$ against the noisy target $y$ is fundamentally misguided. This geometric mismatch exposes the standard inversion process to what we term the \textbf{``Double Noise Trap"}. The optimization objective expands as follows:
\begin{equation}
\begin{aligned}
    \|y_{\mathrm{tar}}-y_{\mathrm{hyp}}\|_2^2
    &=
    \|z-f_{\mathrm{clean}}(\hat{x})\|_2^2 \\
    &\quad
    +2\left\langle z-f_{\mathrm{clean}}(\hat{x}), \epsilon_1-\epsilon_2 \right\rangle
    +\|\epsilon_1-\epsilon_2\|_2^2 .
\end{aligned}
\label{eq:noisy_matching}
\end{equation}
The first term is the desired semantic matching objective, while the remaining terms are caused by noise. As refinement improves and the semantic residual $\|z-f_{\mathrm{clean}}(\hat{x})\|_2^2$ becomes smaller, the stochastic cross term and the doubled noise baseline can dominate the correction signal. The corrector may therefore reward hypotheses whose embeddings align with residual noise directions rather than true semantic directions.

This motivates our decoupled design, which first estimates a denoised clean-manifold representation and then performs generation and correction through denoised embeddings.

\subsection{Infeasible Supervised Denoising}
\label{sec:infeasible_supervised}
The strict no-clean-embedding constraint fundamentally changes the nature of the denoising process. The standard supervised denoising objective,
\begin{equation}
    \min_{\phi}\; \mathbb{E}\left[\|h_{\phi}(y) - z\|_2^2\right],
    \label{eq:oracle_denoising}
\end{equation}
where $h_{\phi}$ denotes a denoising neural network parameterized by $\phi$, is inherently unavailable because the clean target $z$ is unavailable. 

Consequently, these analyzes underscore the critical need for a novel inversion pipeline: one capable of reconstructing the original text relying strictly on the noisy observations.

\section{method}
\label{sec:method}

\begin{figure*}[t]
    \centering
    \includegraphics[page=2,width=\linewidth,
    trim=22 90 10 146,
    clip]{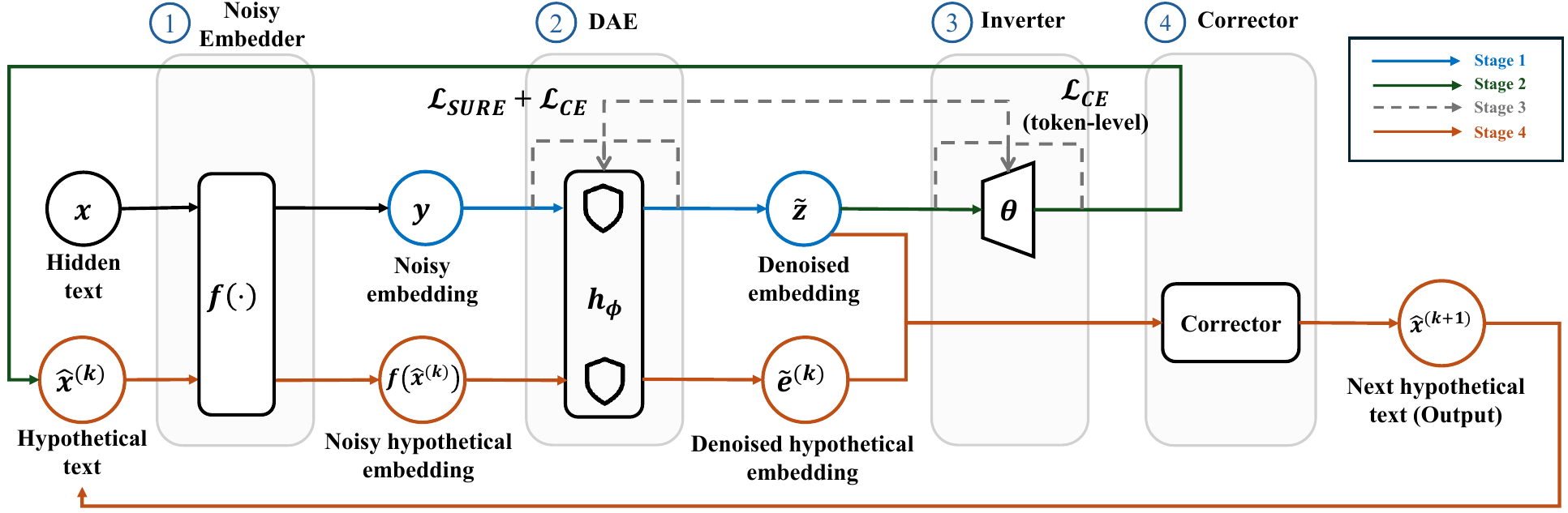}
    \caption{The overall architecture and data flow of our proposed \textsc{DAEI} pipeline. The system consists of four functional components (grey vertical panels) optimized across four stages. (1) Upon obtaining a noisy embedding from the hidden text via the Noisy Embedder, we first apply the DAE (indicated by the shield icons) to reconstruct a denoised embedding $\tilde{z}$. (2) It is then processed by the Inverter to generate an initial hypothetical text $\hat{x}^{(k)}$. (3) During training, the token-level cross-entropy loss ($\mathcal{L}_{\mathrm{CE}}$) backpropagates directly through the active DAE to achieve joint fine-tuning (grey dashed lines). (4) Finally, the generated text enters an iterative correction loop: it is re-encoded by the embedder and passed through the DAE to yield a denoised hypothesis embedding $\tilde{e}^{(k)}$. Both $\tilde{e}^{(k)}$ and the original denoised target $\tilde{z}$ are then fed into the Corrector to predict the next hypothetical text, repeating until the loop terminates to produce the final output.}
    \label{fig:method}
\end{figure*}

We propose \textsc{DAEI}, a novel pipeline for robust noisy text embedding inversion, with its overall architecture and data flow illustrated in Fig. \ref{fig:method}. Our system is built upon four components: a black-box noisy embedder $f(\cdot)$, a denoising autoencoder (DAE) $h_\phi$, a generative inverter $\theta$, and an iterative corrector. Rather than a naive end-to-end approach, the \textsc{DAEI} pipeline comprises four distinct stages: 
(1) SURE-based unsupervised DAE pre-training, 
(2) DAE-shielded inverter training, 
(3) joint \textsc{SURE-CE} fine-tuning, and 
(4) DAE-shielded corrector training. 
Notably, the architectures for Stages 2 and 4 remain virtually identical to vanilla Vec2Text, with the distinction that the input embeddings are denoised by our DAE in advance.

\subsection{SURE-based Unsupervised DAE}
\label{sec:method_sure_dae}

Stage 1 aims to learn a denoising module $h_\phi$ that constructs a denoised embedding $\tilde{z}$ from the noisy observation $y$:
\begin{equation}
    \tilde{z} = h_\phi(y,\sigma).
\end{equation}
The oracle denoising risk  Eq.~(\ref{eq:oracle_denoising}) cannot be directly optimized because the clean embedding $z$ is unobserved. To check if the dependency on $z$ can be removed, we first expand Eq.~(\ref{eq:oracle_denoising}):
\begin{equation}
    \|h_{\phi}(y,\sigma)-z\|_2^2
    =
    \|h_{\phi}(y,\sigma)\|_2^2
    -2 z^\top h_{\phi}(y,\sigma)
    +\|z\|_2^2 .
    \label{eq:oracle_expand}
\end{equation}
The last term $\|z\|_2^2$ is independent of $\phi$, while the cross term $z^\top h_{\phi}(y,\sigma)$ is inaccessible because $z$ is hidden. However, under the Gaussian noise model $y=z+\epsilon$, where $\epsilon\sim\mathcal{N}(0,\sigma^2I_d)$, Stein's lemma gives \cite{stein1981estimation}
\begin{equation}
    \mathbb{E}_{\epsilon}\left[\epsilon^\top h_{\phi}(y,\sigma)\right]
    =
    \sigma^2
    \mathbb{E}_{\epsilon}\left[
    \operatorname{div}_{y} h_{\phi}(y,\sigma)
    \right].
    \label{eq:stein_identity}
\end{equation}
Since $z=y-\epsilon$, the inaccessible cross term can be replaced in expectation by quantities depending only on $y$, the noise variance $\sigma^2$, and the divergence of the denoiser. This leads to SURE \cite{stein1981estimation}:
\begin{equation}
    \mathcal{L}_{\mathrm{SURE}}(y;\phi)
    =
    \|h_{\phi}(y,\sigma)-y\|_2^2
    +2\sigma^2 \operatorname{div}_{y} h_{\phi}(y,\sigma)
    -d\sigma^2.
    \label{eq:sure_general}
\end{equation}
Here, $\operatorname{div}_{y} h_{\phi}(y,\sigma)=\operatorname{Tr}(\partial h_{\phi}/\partial y)$ measures the local sensitivity of the denoiser with respect to its noisy input. Therefore, although $z$ is never observed, minimizing Eq.~(\ref{eq:sure_general}) is equivalent in expectation to minimizing the oracle denoising risk in Eq.~(\ref{eq:oracle_denoising}).

We further parameterize the denoising module as a residual correction around the observed noisy embedding, rather than directly predicting $\tilde{z}$ from scratch:
\begin{equation}
    h_\phi(y,\sigma)= y + r_\phi(y,\sigma),
\end{equation}
where $r_\phi(y,\sigma)$ estimates the displacement needed to suppress the noise component and move the observation back toward the clean embedding manifold. 

Substituting this parameterization into the divergence term gives
$\operatorname{div}_{y}h_{\phi}(y,\sigma)=d+\operatorname{div}_{y}r_{\phi}(y,\sigma)$. After removing constants independent of $\phi$, the training objective becomes
\begin{equation}
    \mathcal{L}_{\mathrm{SURE}}(y;\phi)
    \equiv
    \|r_{\phi}(y,\sigma)\|_2^2
    +2\sigma^2 \operatorname{div}_{y} r_{\phi}(y,\sigma).
    \label{eq:sure_residual}
\end{equation}

We approximate the divergence term with Monte-Carlo Hutchinson estimation \cite{ramani2008monte}:
\begin{equation}
    \operatorname{div}_{y} r_{\phi}(y,\sigma)
    \approx
    \frac{1}{K}\sum_{k=1}^{K}
    v_k^{\top}J_{r_{\phi}}(y,\sigma)v_k,
    \qquad
    v_k \in \{-1,+1\}^{d},
    \label{eq:mc_sure}
\end{equation}
where $J_{r_{\phi}}$ is the Jacobian of the residual network with respect to the noisy embedding. Let $\widehat{\operatorname{div}}_{K}$ denote the Monte-Carlo estimator in Eq.~(\ref{eq:mc_sure}). For Rademacher probes satisfying $\mathbb{E}[v_kv_k^\top]=I$, this estimator is unbiased and its variance is bounded by
\begin{equation}
\begin{aligned}
    \mathbb{E}_{v}
    \left[\widehat{\operatorname{div}}_{K}\right]
    &=
    \operatorname{div}_{y} r_{\phi}(y,\sigma), \\
    \operatorname{Var}
    \left[\widehat{\operatorname{div}}_{K}\right]
    &\leq
    \frac{2}{K}
    \|J_{r_{\phi}}(y,\sigma)\|_F^2 .
\end{aligned}
\label{eq:hutchinson_bound}
\end{equation}
Thus, the approximation is theoretically well grounded: it is unbiased, and its variance decays at rate $O(1/K)$. In practice, we use Rademacher probes and spectral normalization in the residual MLP to stabilize the trace estimate.

\subsection{DAE-Shielded Inverter}
\label{sec:method_dae_first}

After we obtain denoised representation $\tilde{z}$, the inverter $\theta$ is subsequently trained on the pairs $(\tilde{z}, x)$ rather than the raw noisy embeddings. 
It prevents the generative model from overburdening its capacity with high-dimensional noise-filtering and avoids overfitting to non-semantic perturbations.

The training objective is the standard token-level cross-entropy loss:
\begin{equation}
    \mathcal{L}_{\mathrm{CE}}(\theta; \phi) = -\sum_{t=1}^{T} \log p_{\theta}\left(x_t \mid x_{<t}, \tilde{z}\right)
    \label{eq:stage2_ce}
\end{equation}
where $T$ denotes the sequence length, $x_t$ is the target token at step $t$, and $x_{<t}$ represents the preceding autoregressive context.

\subsection{Joint SURE-CE Fine-tuning}
\label{sec:method_joint}

Following the independent training of the DAE and the inverter, stage 3 unifies the architecture. To explicitly align the unsupervised denoising process with the downstream generative task, we jointly fine-tune the DAE and the inverter:

\begin{equation}
    \mathcal{L}_{\mathrm{SURE\mbox{-}CE}}(\phi,\theta)
    =
    \mathcal{L}_{\mathrm{SURE}}(y;\phi)
    +
    \lambda_{\mathrm{CE}}(t)\,
    \mathcal{L}_{\mathrm{CE}}\left(\theta;\phi\right).
    \label{eq:sure_ce}
\end{equation}

Here, the weighting coefficient $\lambda_{\mathrm{CE}}(t)$ is warmed up during the early training steps. The key difference from Stage 2 is that the cross-entropy gradient is allowed to pass through $h_{\phi}(y,\sigma)$. 
Consequently, the DAE receives two complementary supervisory signals: the SURE objective ensures that the output remains strictly aligned with the denoising manifold, while the CE (cross-entropy) loss provides semantic feedback, preserves the latent directions for decoding. 
Concurrently, the inverter is continuously trained on the embeddings denoised dynamically by the fine-tuning DAE to adapt to the optimized embedding distribution.

However, the optimization directions of these two objectives may conflict. We therefore employ distinct learning rates for the DAE and the inverter, and apply gradient surgery PCGrad \cite{yu2020gradient} exclusively to the DAE parameters $\phi$. 

Let $g_{\mathrm{SURE}} \triangleq \nabla_{\phi}\mathcal{L}_{\mathrm{SURE}}$ and $g_{\mathrm{CE}} \triangleq \nabla_{\phi}\mathcal{L}_{\mathrm{CE}}$ denote the gradients with respect to the DAE parameters. If these two gradients exhibit a negative inner product (i.e., $\langle g_{\mathrm{CE}}, g_{\mathrm{SURE}}\rangle < 0$), they are conflicting. In such cases, we project the CE gradient onto the normal plane of the SURE gradient before merging the updates:
\begin{equation}
\begin{aligned}
g_{\mathrm{CE}}'
&=
g_{\mathrm{CE}}
-
\frac{\langle g_{\mathrm{CE}}, g_{\mathrm{SURE}}\rangle}
     {\|g_{\mathrm{SURE}}\|_2^2}
g_{\mathrm{SURE}},
\\
\phi
&\leftarrow
\phi - \eta_{\mathrm{DAE}}
(g_{\mathrm{SURE}} + g_{\mathrm{CE}}').
\end{aligned}
\label{eq:pcgrad}
\end{equation}
This orthogonal projection guarantees that the text feedback ($g_{\mathrm{CE}}'$) shapes the final representation without unlearning the fundamental denoising capabilities.

\subsection{DAE-Shielded Corrector}
\label{sec:stage4_corrector}

Conventionally, once an inverter is trained to generate an initial text hypothesis, a corrector iteratively refines this hypothesis by comparing its embedding against the target embedding. However, when the available target is noisy, directly performing this comparison inevitably triggers the ``Double Noise Trap" described in Section \ref{sec:dn_trap}. 

To mitigate this, we introduce a \textit{DAE Shield} to both sides of the correction interface. Specifically, instead of comparing raw representations, we pass both the noisy target $y$ and the embedding of the current hypothesis $\hat{x}^{(k)}$ through our Stage 3 denoiser $h_{\phi^\star}$:
\begin{equation}
    \tilde{z} = h_{\phi^\star}(y, \sigma),
    \qquad
    \tilde{e}^{(k)} = h_{\phi^\star}(f(\hat{x}^{(k)}), \sigma),
    \label{eq:dae_shield}
\end{equation}
where $\hat{x}^{(k)}$ is the generated text at correction step $k$, and $\phi^\star$ denotes the fully fine-tuned DAE parameters. The corrector then conditions exclusively on the denoised target $\tilde{z}$, the denoised hypothesis $\tilde{e}^{(k)}$, and their residual difference. By completely shielding the correction interface, we constrain the iterative refinement strictly within the learned clean manifold, ensuring the model aligns with semantic text directions rather than chasing irreducible noise residuals.

\section{Experimental Setup}

Our study aims to determine the extent to which noise-protected embeddings from black-box encoders remain vulnerable to advanced inversion attack paradigms. Specifically, we investigate the following research questions: 
\textbf{RQ1:} How does the proposed \textsc{DAEI} pipeline compare against vanilla Vec2Text and existing noise-robust baselines in recovering target text from noisy embeddings?
\textbf{RQ2:} Can the proposed \textsc{DAEI} pipeline be utilized in different embedding model backbones?
\textbf{RQ3:} Can \textsc{DAEI} maintain strong inversion performance on out-of-domain test sets? 
\textbf{RQ4:} Can the DAE component effectively reduce noise?
\textbf{RQ5:} Does the joint fine-tuning module improve reconstruction quality?
\textbf{RQ6:} To what extent can \textsc{DAEI} maintain its inversion performance when the noise ${\sigma}$ varies? 

\subsection{Dataset}

\subsubsection{Training set}

All main experiments use the same training mixture of:
\begin{itemize}
    \item \textbf{Natural Questions (NQ) } is a question answering benchmark built from real Google search queries \cite{kwiatkowski2019natural}. 

    \item \textbf{MS MARCO} is a large-scale machine reading retrieval dataset constructed from real Bing queries, web passages, and human-generated answers \cite{bajaj2016ms}. 

    \item \textbf{Yahoo Answers} is a community question-answering domain covering broad everyday topics. 
\end{itemize}
We capped the combined mixture at 7M text examples as raw text sources to construct the inversion datasets. While the underlying text corpora remain constant, the exact input-target configurations vary across our four training stages to accommodate the specific objectives of each module:

\begin{itemize}
    \item \textbf{Stage 1 (DAE Pre-training):} To train the residual DAE, the system requires \textit{noisy embeddings} only.
    \item \textbf{Stage 2 (Base Inverter Training):} To train the initial text decoding module, we utilize \textit{(denoised embedding, text)} pairs. These denoised embeddings are computed by the Stage 1 DAE.
    \item \textbf{Stage 3 (Joint Fine-tuning):} For end-to-end joint optimization, the dataset consists of \textit{(noisy embedding, text)} pairs. Here, the noisy embeddings are dynamically denoised on-the-fly by the actively tuning DAE during the forward pass.
    \item \textbf{Stage 4 (Corrector Training):} To train the iterative corrector, we rely on expanded input tuples consisting of \textit{(denoised embedding, hypothetical text, denoised hypothetical embedding)}, paired with the original text as label. Crucially, these denoised embeddings are computed by the fully tuned Stage 3 DAE, ensuring the corrector operates strictly within the optimal denoised distribution.
\end{itemize}

\subsubsection{Evaluation sets}
For evaluation, we consider both in-domain and out-of-domain settings. The \textbf{in-domain} test set is the held-out splits from the same NQ, MS MARCO, and Yahoo Answers mixture, while the \textbf{out-of-domain} test sets are drawn from five unseen corpora: AG News, Anthropic Toxic Prompts, Python Code Alpaca, Climate-FEVER \cite{diggelmann2020climate}, and MedMCQA \cite{pal2022medmcqa}. These out-of-domain datasets allow us to evaluate whether the inversion methods generalize beyond the training distribution. The data sources for all training and evaluation sets are provided in our repository: https://anonymous.4open.science/r/DAEI-1F1C/


\subsection{Target Embedding Models}
To rigorously evaluate the generalizability of our \textsc{DAEI} pipeline across different latent space distributions, we select two widely adopted, state-of-the-art dense text embedding models as our black-box targets:

\begin{itemize}
    \item \textbf{GTR-base} \cite{ni2022large}: The GTR (Generalizable T5-based dense Retrievers) model leverages a bi-encoder architecture initialized from the T5 \cite{raffel2020exploring} checkpoints. It is specifically optimized for robust zero-shot retrieval tasks, producing highly structured semantic representations.
    \item \textbf{GTE-base} \cite{li2023towards}: The General Text Embeddings (GTE) model trained on large-scale contrastive data represents a highly optimized encoder-only architecture. 
\end{itemize}

 We deliberately select GTR and GTE since they represent two fundamentally distinct backbone architectures (T5-based vs. BERT-based) and training paradigms, thereby demonstrate the generalizability of our method.

\subsection{Baselines and Our Models}
 Given the scarcity of existing work on noisy text embedding inversion, we use \textsc{ZSInvert} \cite{zhang2025universal} (an adversarial decoding approach) as our primary external baseline even though it is not based on the mainstream encoder-decoder architecture because it is the only existing work that we know has shown its effectiveness in the noisy text embedding inversion scenario. Our core focus remains on a rigorous comparison between various configurations of the encoder-decoder models and our proposed approach. To comprehensively assess inversion performance, we categorize the baselines and our models into three distinct groups:

\subsubsection{Clean Upper Bound}  \textbf{\textsc{Clean-to-Clean} Baseline} evaluates the vanilla Vec2Text \cite{morris2023text} model on clean embeddings. It represents the state-of-the-art text recovery performance under ideal, noiseless conditions. This serves as the theoretical ceiling that our noisy inversion pipelines aim to approach.

\subsubsection{Generative Inversion under Noise} This group compares the standard autoregressive inversion models against our proposed pipelines when facing perturbed embeddings:
\begin{itemize}
    \item \textbf{\textsc{Clean-to-Noisy} Baseline:} Tests the severe distribution shift faced by a vanilla Vec2Text model that is trained solely on clean embeddings but evaluated on noisy ones.
    \item \textbf{\textsc{Noisy Inverter} Baseline:} Trains a vanilla Vec2Text directly on noisy embeddings. This provides a much stronger baseline that adapts to the corrupted embedding distribution but lacks denoising mechanism.
    \item \textbf{\textsc{DAE-only} inverter (Ours):} Our proposed pipeline without joint fine-tuning. A DAE first denoises the embedding, then an inverter is trained on the denoised cache.
    \item \textbf{\textsc{DAEI} (Ours):} Our complete pipeline that jointly fine-tunes both the DAE and the inverter. 
\end{itemize}

\subsubsection{Adversarial-Decoding-Based Baseline} To step outside the generative paradigm, we introduce \textsc{ZSInvert} as an external baseline. This method frames inversion as adversarial decoding, iteratively optimizing discrete tokens to minimize embedding distance. 
    
Table~\ref{tab:exp_systems} summarizes the above systems evaluated in our main experiments.

\begin{table}[t]
\centering
\caption{Main experimental systems. \textsc{ZSInvert} base is training-free; its listed training embedding type refers only to corrector training.}
\label{tab:exp_systems}
\renewcommand{\arraystretch}{1.2} 
\resizebox{\columnwidth}{!}{%
\begin{tabular}{@{}lccc@{}} 
\toprule 
System & Train embedding & Test embedding  \\
\midrule 
\textsc{Clean-to-Clean} Baseline & Clean & Clean  \\
\textsc{ZSInvert} & Noisy & Noisy \\
\textsc{Clean-to-Noisy} Baseline & Clean & Noisy \\
\textsc{Noisy Inverter} Baseline & Noisy & Noisy \\
\textsc{DAE-only} (ours) & Denoised by stage1 DAE & Denoised by stage1 DAE \\
\textsc{DAEI} (ours) & Denoised by stage3 DAE & Denoised by stage3 DAE \\

\bottomrule 
\end{tabular}
}
\end{table}

\subsection{Metrics.}
For evaluating the inverters, we report sentence-level BLEU \cite{papineni2002bleu}, token-level F1 \cite{rajpurkar2016squad}, ROUGE-L \cite{lin2004rouge}, and the cosine similarity between the embeddings of the reconstructed and reference texts. Specifically, BLEU assesses generation precision and local fluency by measuring the n-gram overlap between the reconstructed and reference texts; token-level F1 assesses keyword recovery by measuring unigram overlap independently of sequence order; ROUGE-L evaluates structural similarity and information recall based on the longest common subsequence; and embedding cosine similarity measures semantic closeness in the target embedding space.

For evaluating the DAE, we also report the MSE and embedding cosine similarity between the DAE's denoised outputs and the clean embeddings. We emphasize that these clean embeddings are strictly isolated for offline evaluation purposes only; they are never observed by the model during any stage of training, nor are they used to compute any optimization loss.

\subsection{Implementation Setting.}
Our method assumes the noise scale $\sigma$ is prior knowledge (see Section \ref{sec:problem_task}). Unless otherwise specified in the ablation studies, we apply zero-mean Gaussian noise with a standard deviation of $\sigma = 0.01$ \cite{morris2023text,zhuang2024understanding}. The DAE module is pre-trained with a learning rate of $1 \times 10^{-3}$. We configure a hidden dimension of 1024 and a depth of 3, spectral normalization, and five Monte-Carlo probes.

All inverter variants use the Vec2Text architecture with a T5 \cite{raffel2020exploring} backbone and are optimized with a learning rate of $1 \times 10^{-3}$. Following the original Vec2Text setting \cite{morris2023text}, both input and target texts are truncated to a maximum length of 128 tokens during training, while evaluation is conducted with a maximum length of 32 tokens. 

During the joint fine-tuning stage, the DAE learning rate is reduced to $2 \times 10^{-4}$, and the cross-entropy loss weight is linearly warmed up over 10,000 steps. 

All corrector variants use the Vec2Text architecture with a learning rate of $5 \times 10^{-4}$. We enforce an early-stop threshold of $1 \times 10^{-3}$: refinement terminates once the cosine similarity gain between consecutive steps falls below this value, preventing the corrector from overfitting to residual noise.

\section{Results}
\label{sec:results}

\begin{table*}[htbp]
    \centering
    \caption{Main results for in-domain evaluation with \textbf{GTR-base}. Gray rows indicate reference settings that are \textbf{not} available under the current threat model. Improvements in parentheses are absolute gains over the \textbf{\textsc{Noisy Inverter} Baseline}. Best feasible results excluding the clean upper bound are in \textbf{bold}.}
    \label{tab:main_results_gtr}
    \begin{tabular}{llcccc}
        \toprule
        \textbf{Category} & \textbf{Model} & \textbf{BLEU} & \textbf{F1} & \textbf{ROUGE-L} & \textbf{Cosine Similarity} \\
        \midrule
        Clean Upper Bound 
        & \textsc{Clean-to-Clean} Baseline
        & \textcolor{gray}{0.6252} 
        & \textcolor{gray}{0.8656} 
        & \textcolor{gray}{0.8704} 
        & \textcolor{gray}{0.9500} \\
        
        \midrule
        Adversarial Decoding
        & \textsc{ZSInvert} 
        & 0.013 & 0.1503 & 0.1117 & 0.8568 \\
        
        \midrule
        \multirow{4}{*}{Generative Inversion}
        & \textsc{Clean-to-Noisy} Baseline
        & \textcolor{gray}{0.1348} 
        & \textcolor{gray}{0.5005} 
        & \textcolor{gray}{0.5012} 
        & \textcolor{gray}{0.7785} \\
        & \textsc{Noisy Inverter} Baseline 
        & 0.1916 & 0.5814 & 0.5660 & 0.8349 \\
        & \textsc{DAE-only} (Ours) 
        & 0.3616 & 0.7206 & 0.7200 & 0.9055 \\
        & \textsc{DAEI} (Ours) 
        & \textbf{0.4875} (+0.2959) 
        & \textbf{0.7717} (+0.1903) 
        & \textbf{0.7788} (+0.2128) 
        & \textbf{0.9254} (+0.0905) \\
        \bottomrule
    \end{tabular}
\end{table*}

\begin{table*}[htbp]
    \centering
    \caption{Main results for in-domain evaluation with \textbf{GTE-base}. Formatting follows Table~\ref{tab:main_results_gtr}.}
    \label{tab:main_results_gte}
    \begin{tabular}{llcccc}
        \toprule
        \textbf{Category} & \textbf{Model} & \textbf{BLEU} & \textbf{F1} & \textbf{ROUGE-L} & \textbf{Cosine Similarity} \\
        \midrule
        Clean Upper Bound 
        & \textsc{Clean-to-Clean} Baseline
        & \textcolor{gray}{0.4391} 
        & \textcolor{gray}{0.7622} 
        & \textcolor{gray}{0.7570} 
        & \textcolor{gray}{0.9211} \\
        
        \midrule
        Adversarial Decoding
        & \textsc{ZSInvert} & 0.0104 & 0.1189 & 0.0894 & \textbf{0.9641} \\
        
        \midrule
        \multirow{4}{*}{Generative Inversion}
        & \textsc{Clean-to-Noisy} Baseline
        & \textcolor{gray}{0.0846} 
        & \textcolor{gray}{0.3222} 
        & \textcolor{gray}{0.3271} 
        & \textcolor{gray}{0.7081} \\
        & \textsc{Noisy Inverter} Baseline 
        & 0.1249 & 0.3960 & 0.4044 & 0.7520 \\
        & \textsc{DAE-only} (Ours) 
        & 0.2688 & 0.5854 & 0.5722 & 0.8141 \\
        & \textsc{DAEI} (Ours) 
        & \textbf{0.3152} (+0.1903) 
        & \textbf{0.6344} (+0.2384) 
        & \textbf{0.6324} (+0.2280) 
        & 0.8556 (+0.1036) \\
        \bottomrule
    \end{tabular}
\end{table*}

\subsection{Overall Performance (RQ1 \& RQ2)}
\textbf{\textsc{DAEI} achieves the best noisy inversion performance.}
Tables \ref{tab:main_results_gtr} and \ref{tab:main_results_gte} report the main inversion results on GTR-base and GTE-base, respectively. On GTR-base, \textsc{DAEI} substantially improves over the \textsc{Noisy Inverter} baselines across all metrics. In particular, \textsc{DAEI} increases BLEU from 0.1916 to 0.4875, yielding an absolute gain of nearly 30 percentage points and a 154.4\% relative improvement over the \textsc{Noisy Inverter} baseline. It also improves token-level F1 from 0.5814 to 0.7717 and ROUGE-L from 0.5660 to 0.7788, indicating that \textsc{DAEI} can recover richer textual information from noisy embeddings alone. Moreover, the embedding cosine similarity increases from 0.8349 to 0.9254, suggesting that the DAE-shielded correction process more effectively aligns the hypothesis embedding with the target embedding after removing a substantial portion of the perturbation noise.

\textbf{Noisy training alone provides limited gains.}
The comparison between the \textsc{Clean-to-Noisy} baseline and the \textsc{Noisy Inverter} baseline further illustrates the limitation of directly training Vec2Text on noisy embedding. On GTR-base, noisy training improves text reconstruction metrics by only 5--8 points over the \textsc{Clean-to-Noisy} baseline, indicating limited 
adaptation to noisy inputs. However, due to the Double Noise Trap, the correction stage compares a noisy hypothesis embedding with a separately perturbed target, causing the residual signal to be contaminated by doubled noise. As a result, this gain remains far lower than that achieved by \textsc{DAEI}.

\textbf{\textsc{DAEI} approaches the clean upper bound and outperforms \textsc{ZSInvert} in text reconstruction.}
Relative to the clean upper bound, \textsc{DAEI} recovers 78.0\%/89.2\% of BLEU/F1 on GTR-base and 71.7\%/83.2\% on GTE-base. These results suggest that noise-protected embeddings still retain considerable recoverable textual information when the adversary adopts a denoising-aware inversion pipeline. However, \textsc{ZSInvert} performs poorly on text-overlap metrics for both embedders, achieving only around 0.01 BLEU and 0.08--0.15 on F1 and ROUGE-L, despite obtaining a relatively high embedding cosine similarity of 0.85--0.96. 
This gap indicates that adversarial decoding matches the embedding space but fails to reliably recover token-aligned text, whereas \textsc{DAEI} better balances semantic matching and fluent reconstruction.

\textbf{\textsc{DAEI} generalizes across embedding backbones.}
\textsc{DAEI} shows similar gains on GTE-base, answering RQ2. Although the overall scores on GTE-base are lower than GTR-base, the clean upper bound is also lower, suggesting that the underlying embedding space is intrinsically more difficult for Vec2Text framework to invert. One possible reason is that our inverter uses a T5-base backbone, which is more closely aligned with the T5-based GTR encoder than with the BERT-style GTE encoder \cite{li2023towards,ni2022large}; meanwhile, GTE is trained with multi-stage contrastive learning for general-purpose semantic retrieval \cite{li2023towards}, which may emphasize semantic invariance while preserving less information needed for token-level reconstruction. 
Even under this harder setting, \textsc{DAEI} improves over the \textsc{Noisy Inverter} baseline by approximately 19--23 points on text reconstruction metrics, while also increasing cosine similarity by 10.36 points. 
These consistent improvements over two architecturally different encoders demonstrate that \textsc{DAEI} is generalized across different noisy embedding interfaces.

\begin{table}[htbp]
    \centering
    \caption{Out-of-domain evaluation results for \textsc{DAEI}. Upper Bound denotes \textsc{Clean-to-Clean} Baseline, and Baseline denotes the \textsc{Noisy Inverter} Baseline.}
    \label{tab:ood_results}
    \begin{tabular}{llccc}
        \toprule
        \textbf{OOD Dataset} & \textbf{Method} & \textbf{BLEU} & \textbf{F1} & \textbf{ROUGE-L} \\
        \midrule
        \multirow{3}{*}{AG News}
            & Upper Bound & \textcolor{gray}{0.1023} & \textcolor{gray}{0.4296} & \textcolor{gray}{0.4887} \\
            & \textsc{ZSInvert} & 0.0130 & 0.1629 & 0.1142 \\
            & Baseline & 0.0531 & 0.2749 & 0.3109 \\
            & \textsc{DAEI} & \textbf{0.0791} & \textbf{0.3346} & \textbf{0.3915} \\
        \midrule
        \multirow{3}{*}{\begin{tabular}[c]{@{}l@{}}Anthropic\\ Toxic Prompts\end{tabular}}
            & Upper Bound & \textcolor{gray}{0.8114} & \textcolor{gray}{0.8912} & \textcolor{gray}{0.9118} \\
            & \textsc{ZSInvert} & 0.0137 & 0.1451 & 0.1149 \\
            & Baseline & 0.2128 & 0.5581 & 0.5590 \\
            & \textsc{DAEI} & \textbf{0.7217} & \textbf{0.8881} & \textbf{0.8885} \\
        \midrule
        \multirow{3}{*}{Python Code}
            & Upper Bound & \textcolor{gray}{0.6732} & \textcolor{gray}{0.8588} & \textcolor{gray}{0.8783} \\
            & \textsc{ZSInvert} & 0.0174 & 0.1928 & 0.1434 \\            
            & Baseline & 0.1933 & 0.6085 & 0.5961 \\
            & \textsc{DAEI} & \textbf{0.5080} & \textbf{0.8213} & \textbf{0.8316} \\
        \midrule
        \multirow{3}{*}{Climate fever}
            & Upper Bound & \textcolor{gray}{0.5350} & \textcolor{gray}{0.7843} & \textcolor{gray}{0.7824} \\
            & \textsc{ZSInvert} & 0.0132 & 0.1518 & 0.1105 \\
            & Baseline & 0.1377 & 0.4853 & 0.4980 \\
            & \textsc{DAEI} & \textbf{0.3489} & \textbf{0.6651} & \textbf{0.6961} \\
        \midrule
        \multirow{3}{*}{Medmcqa}
            & Upper Bound & \textcolor{gray}{0.6689} & \textcolor{gray}{0.8476} & \textcolor{gray}{0.8800} \\
            & \textsc{ZSInvert} & 0.0128 & 0.1286 & 0.1070 \\
            & Baseline & 0.1739 & 0.5252 & 0.5576 \\
            & \textsc{DAEI} & \textbf{0.4871} & \textbf{0.7730} & \textbf{0.8037} \\
        \bottomrule
    \end{tabular}
\end{table}

\subsection{Out-of-Domain Generalization (RQ3)}
Table~\ref{tab:ood_results} evaluates whether \textsc{DAEI} can generalize to Out-Of-Domain (OOD) test sets. Across all OOD datasets, \textsc{DAEI} consistently outperforms the \textsc{Noisy Inverter} and \textsc{ZSInvert} on BLEU, F1, and ROUGE-L, showing that \textsc{DAEI} remains effective when the target texts come from unseen domains.

The performance varies across datasets. \textsc{DAEI} performs especially well on Anthropic Toxic Prompts, Python Code, and MedMCQA, while AG News obtains lower scores for both the clean upper bound and \textsc{DAEI}. One likely reason is that AG News examples are longer on average, with about 61 tokens compared with 22--30 tokens for the other datasets. Although evaluation outputs are truncated to 32 tokens, longer source texts still contain more information to recover and are more likely to suffer from truncation and partial reconstruction. 

\begin{table}[htbp]
    \centering
    \scriptsize
    \caption{DAE component performance across training stages. MSE and cosine similarity are computed against clean embeddings before and after DAE denoising, with $\Delta$ reporting the denoising improvement.} 
    \label{tab:dae_exp}
    \setlength{\tabcolsep}{3pt}
    \renewcommand{\arraystretch}{1.05}
    \resizebox{\columnwidth}{!}{
    \begin{tabular}{ll ccc ccc}
        \toprule
        \multirow{2}{*}{\textbf{Embedder}} & \multirow{2}{*}{\textbf{Stage}} & \multicolumn{3}{c}{\textbf{MSE $\downarrow$}} & \multicolumn{3}{c}{\textbf{Cosine $\uparrow$}} \\
        \cmidrule(lr){3-5} \cmidrule(lr){6-8}
        & & \textbf{Noisy} & \textbf{DAE} & \textbf{$\Delta$} & \textbf{Noisy} & \textbf{DAE} & \textbf{$\Delta$} \\
        \midrule
        \multirow{2}{*}{GTE-base} 
        & Stage 1 & 0.0765 & 0.0583 & 0.0182 & 0.9662 & 0.9727 & 0.0065 \\
        & Stage 3 & 0.0765 & 0.0602 & 0.0163 & 0.9662 & 0.9713 & 0.0051 \\
        \midrule
        \multirow{2}{*}{GTR-base} 
        & Stage 1 & 0.0766 & 0.0542 & 0.0224 & 0.9179 & 0.9322 & 0.0142 \\
        & Stage 3 & 0.0766 & 0.0575 & 0.0191 & 0.9179 & 0.9280 & 0.0101 \\
        \bottomrule
    \end{tabular}
    }
\end{table}

\begin{figure}[htbp]
    \centering
    \begin{subfigure}{0.48\linewidth}
        \centering
        \includegraphics[width=\linewidth]{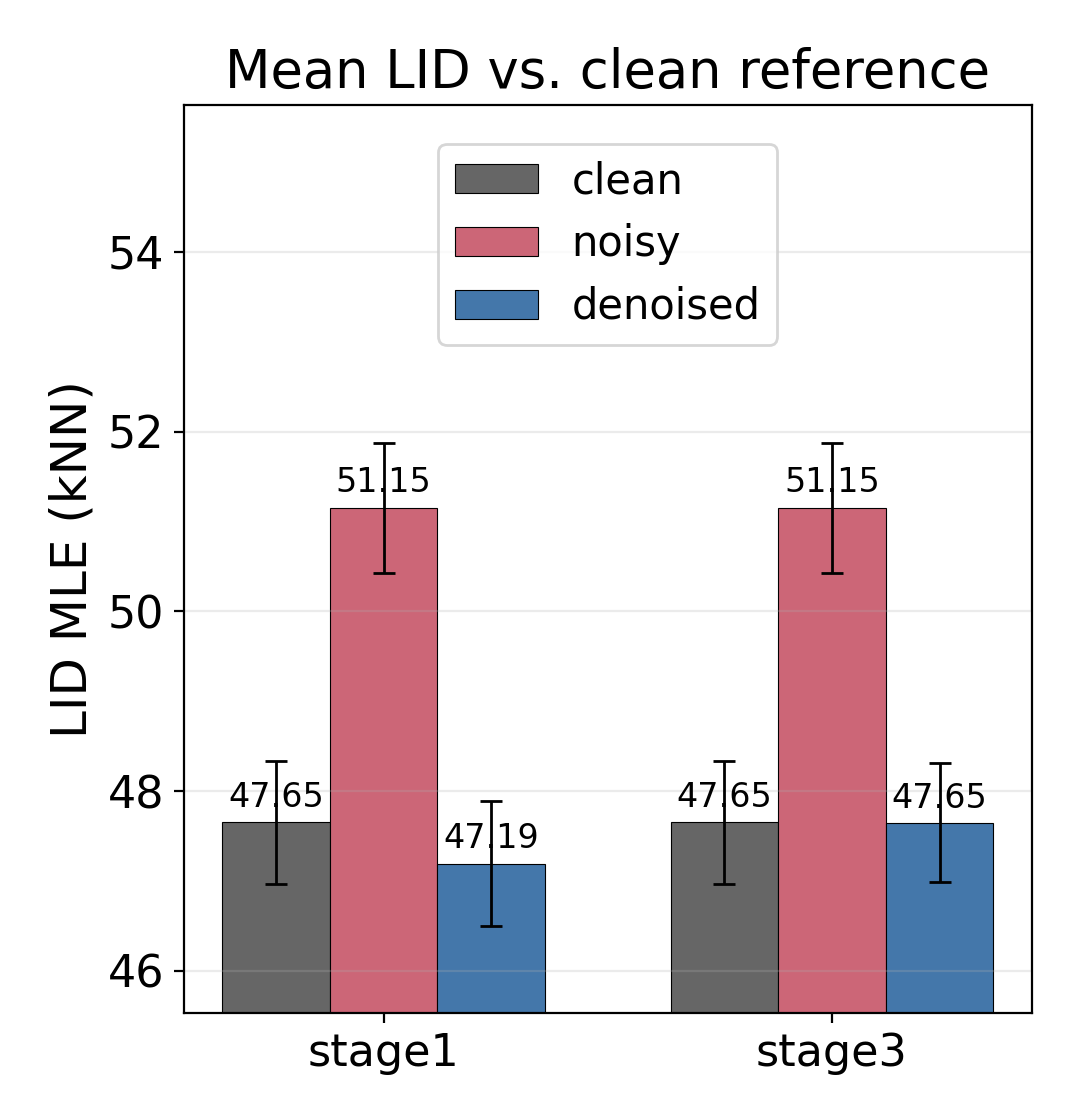}
        \caption{Mean LID with SEM.}
        \label{fig:lid_mean}
    \end{subfigure}
    \hfill
    \begin{subfigure}{0.48\linewidth}
        \centering
        \includegraphics[width=\linewidth]{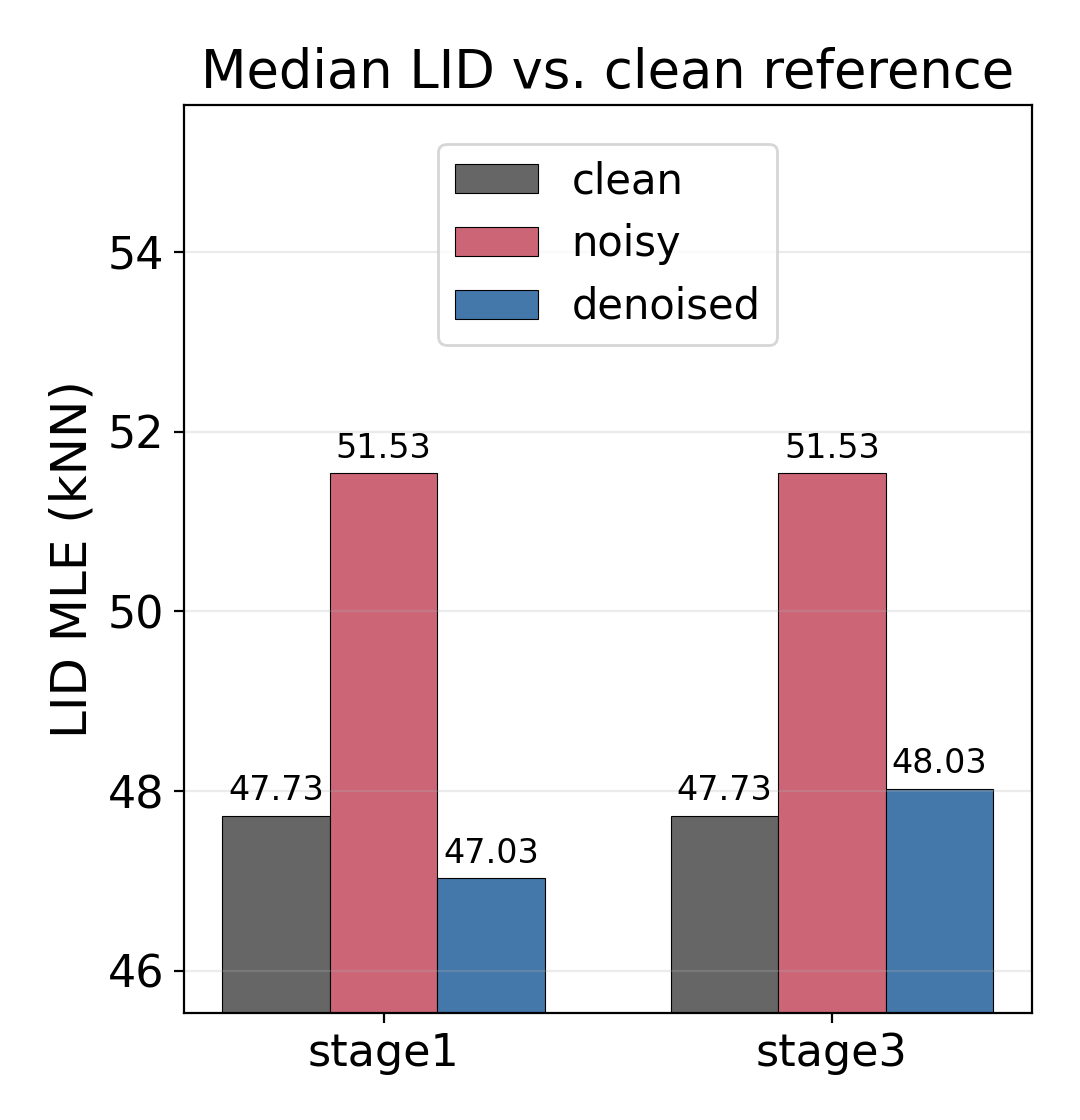}
        \caption{Median LID.}
        \label{fig:lid_median}
    \end{subfigure}
    \caption{LID-based geometric evaluation of DAE projection using clean embeddings as the reference set. The two subfigures report mean LID with SEM (Standard Error of the Mean) error bars and median LID, respectively, showing DAE reduces the LID of noisy embeddings back to the clean level.}
    \label{fig:lid_projection}
\end{figure}

\subsection{Components Effectiveness (RQ4 \& RQ5)}

\subsubsection{DAE Effectiveness}
\textbf{DAE reduces noise-induced embedding distortion.}
Table \ref{tab:dae_exp} demonstrates DAE can effectively reduce the perturbation noise without using clean embeddings during training. The Stage~1 DAE reduces MSE to the clean embeddings by 29.2\% on GTR-base and 23.8\% on GTE-base, while improving cosine similarity from 0.9179 to 0.9322 and from 0.9662 to 0.9727, respectively.
Since the injected noise is relatively small, the raw noisy embeddings are already close to the clean embeddings in cosine similarity, making the cosine gains modest. Nevertheless, the consistent MSE reduction and cosine improvement show that the DAE shifts noisy embeddings closer to their clean counterparts, even without using clean embeddings as training targets.

\textbf{Different roles of Stage 1 and Stage 3 DAEs.}
Interestingly, Table~\ref{tab:dae_exp} shows that the Stage~3 DAE has slightly weaker overall denoising metrics than the Stage~1 DAE in terms of MSE and cosine similarity to the clean embedding. This is expected because the two stages optimize for different objectives: Stage~1 focuses purely on denoising, while Stage~3 is additionally guided by the cross-entropy objective toward semantic directions for text reconstruction, resulting in a representation that is better aligned with the generative decoder.

\textbf{DAE projects noisy embeddings toward the clean manifold.}
We further evaluate the DAE from a geometric perspective using Local Intrinsic Dimensionality (LID) \cite{houle2017local}. For each test embedding, we estimate its LID with respect to a held-out clean embedding reference set using the MLE (Maximum Likelihood Estimation) estimator with 100 nearest neighbors. 
As shown in Fig.~\ref{fig:lid_projection}, Gaussian noise increases the mean LID from 47.65 to 51.15, indicating that noisy embeddings deviate from the clean embedding manifold. After DAE projection, the LID decreases back to the clean level: Stage~1 obtains a slightly lower LID (47.19) than clean embedding, it may because a purely denoising DAE pulls embeddings toward more averaged regions of the clean distribution, result in lower estimated local dimensionality. In contrast, Stage~3 reaches 47.65, almost exactly matching the clean embeddings, because joint fine-tuning introduces semantic alignment with the clean manifold. The median results show the same trend, confirming that the reduction is not driven by outliers.

\textbf{DAE improves downstream inversion.}
The effectiveness of the DAE is also reflected in the downstream inversion results through the comparison between the \textsc{Noisy Inverter} baseline and \textsc{DAE-only}. Across the two embedders, \textsc{DAE-only} improves the text reconstruction metrics by approximately 14--19 percentage points, while increasing embedding cosine similarity by about 6--7 percentage points. This demonstrates the fundamental effectiveness of the denoising-aware strategy: removing part of the perturbation noise before inversion makes the input representation more semantically stable and easier for the generative decoder to interpret. 

\subsubsection{Role of Joint Fine-tuning}

While \textsc{DAE-only} already provides strong gains, the full \textsc{DAEI} model further improves performance by jointly fine-tuning the DAE and the inverter. Across the two embedders, \textsc{DAEI} brings additional improvements of approximately 5--13 percentage points on text reconstruction metrics over \textsc{DAE-only}, while also further increasing embedding cosine similarity by 2--4 points. These consistent gains highlight the effectiveness of joint fine-tuning, showing that a small sacrifice in denoising accuracy can be worthwhile when it better aligns the representation with the downstream text decoding objective.


\begin{table}[htbp]
    \centering
    \caption{In-domain evaluation of \textsc{DAEI} on GTR-base under different Gaussian noise scales.}
    \label{tab:sigma_results}
    \begin{tabular}{ccccc}
        \toprule
        $\sigma$ & \textbf{BLEU} & \textbf{F1} & \textbf{ROUGE-L} & \textbf{Cosine} \\
        \midrule
        0.005 & 0.4927 & \textbf{0.7823} & \textbf{0.7850} & \textbf{0.9293} \\
        0.008 & \textbf{0.4932} & 0.7807 & 0.7836 & 0.9288 \\
        0.010 & 0.4875 & 0.7717 & 0.7788 & 0.9254 \\
        0.012 & 0.4813 & 0.7729 & 0.7793 & 0.9242 \\
        0.015 & 0.4628 & 0.7578 & 0.7634 & 0.9218 \\
        0.018 & 0.4377 & 0.7421 & 0.7480 & 0.9240 \\
        0.020 & 0.3966 & 0.7200 & 0.7269 & 0.9150 \\
        \bottomrule
    \end{tabular}
\end{table}

\subsection{Generalization Across Varying Noise Levels (RQ6)}
Table~\ref{tab:sigma_results} evaluates the in-domain robustness of \textsc{DAEI} on GTR-base under different Gaussian noise scales. The model is trained with $\sigma=0.01$ and tested on other noise levels. \textsc{DAEI} remains highly stable when $\sigma$ varies from 0.005 to 0.018: BLEU stays above 0.4377, F1 remains above 0.7421, and cosine similarity remains around 0.924--0.929. Even at $\sigma=0.02$, \textsc{DAEI} still achieves 0.3966 BLEU and 0.7200 F1, indicating that \textsc{DAEI} can tolerate moderate mismatch between the training noise scale and the test-time perturbation. 
This result is especially relevant when the defender does not use a fixed noise scale, but instead samples $\sigma$ from a small range, such as $[0.005, 0.015]$, for each released embedding: although the attacker may only estimate an average $\sigma$, \textsc{DAEI} can still work effectively. Larger noise scales may further reduce inversion performance, but prior work has shown that overly large perturbations also damage the downstream utility of the protected embedder~\cite{morris2023text}. Thus, within the practical noise regime, \textsc{DAEI} is robust to realistic noise-scale fluctuations.

\section{Related Work}
\label{sec:related}

\subsection{Text Embedding Inversion Attacks} Dense text embeddings can leak substantial information about their original inputs. Early work demonstrated this risk by showing that sentence embeddings can reveal sensitive attributes and keywords \cite{song2020information}. Later, Li et al. proposed a generative embedding inversion attack that reconstructs complete sentences \cite{li2023sentence}. Morris et al. introduced Vec2Text, which achieves high-fidelity reconstruction of short texts \cite{morris2023text}, while later methods reduce attacker requirements via few-shot, transfer-based, or zero-shot inversion~\cite{chen2025algen,huang2024transferable,zhang2025universal}. These works demonstrate the practical privacy risk of text embeddings, but they mainly focus on general embeddings settings rather than embeddings explicitly protected by noise.

\subsection{Privacy Protection for Text Embeddings} Noise injection is a standard mechanism for privacy protection, with differential privacy providing a formal foundation for calibrated random perturbations \cite{dwork2006calibrating}. In NLP, Feyisetan et al. applied calibrated multivariate noise in word embedding space to achieve privacy-utility tradeoffs for textual analysis \cite{feyisetan2020privacy}, and Xu et al. further refined this idea with a Mahalanobis metric \cite{xu2020differentially}. More recent work extends perturbation-based privacy to sentence-level representations \cite{du2023sanitizing,bollegala2025metric}. Recent evaluations also show that Gaussian noise can degrade the performance of standard embedding inversion attacks \cite{seputis2025rethinking,morris2023text,zhang2025universal}. Such a lightweight but effective defense poses a substantial challenge to embedding inversion attacks.

\subsection{Denoising Techniques} Denoising autoencoders learn robust representations by reconstructing clean inputs from corrupted observations\cite{vincent2008extracting}. Standard denoising training typically relies on clean reconstruction targets. However, Noise2Noise relaxes this requirement by learning from pairs of independently corrupted observations of the same underlying signal \cite{lehtinen2018noise2noise}. Stein's unbiased risk estimate provides a way to estimate Gaussian denoising risk from noisy observations alone \cite{stein1981estimation}, and Monte Carlo SURE extends this idea to general denoisers through randomized divergence estimation \cite{ramani2008monte,soltanayev2018training}. However, these techniques have not been studied in the context of text embedding inversion.

As a result, the community still lacks a systematic understanding of whether high-quality text can be reconstructed from noise-protected embeddings. This gap motivates our study of denoising-aware embedding inversion in the noisy-only setting.

\section{Conclusion}
\label{sec:conclusion}

In this paper, we studied the privacy risk of noise-protected text embeddings under a threat model where attackers observe only perturbed embeddings. We propose \textsc{DAEI}, a denoising-aware embedding inversion pipeline that combines unsupervised  denoising autoencoder with generative inversion. Extensive experiments across embedding backbones, domains, and ablation settings show that \textsc{DAEI} outperforms existing baselines, demonstrating that Gaussian noise alone is insufficient to prevent textual information leakage. These findings highlight the need to evaluate privacy-preserving embedding systems against adaptive denoising-aware attacks and to develop stronger defenses for embedding-based applications.

\section{Acknowledgment}
This work was supported by Monash eResearch capabilities, including M3 HPC, the Australian Research Council through the Discovery Early Career Researcher Award (DE250100032), the Discovery Project (Grant No. DP260100326) and the Linkage Projects (Grant Nos. LP230200892, LP240200546 and LP250200778).

\bibliographystyle{IEEEtran}
\bibliography{ref}

\end{document}